# A plant-inspired Multifunctional, Bi-directional, and Fiberless Soft Gripper with Embedded Sensing Ability

Mohsen Annabestani, *Member, IEEE*, Majid Shabani, Samuel Videira Magalhaes, Alessio Mondini, and Barbara Mazzolai, *Member, IEEE*

*Abstract—* This work presents a new fiberless soft pneumatic actuator that can work multifunctional and bidirectional, and its embedded sensors give it a self- proprioception ability. This actuator works based on the idea of employing helical pressure channels. Applying the controlled input pressures into these two channels causes a variety of deformations and actuation. In particular, single pressure, imbalanced pressures, and balanced pressures applied in the channels cause bidirectional coilings, opposite bendings, and elongation, respectively, in a single unit actuator. Also, two U-shaped microchannels are created, and by injecting a gel-based conductive material, the actuator is equipped with resistive sensors which are responsive to a vast dynamic range from a small oscillation to a large elongation. This actuator has so many promising features as a multifunctional soft gripper, and its embedded soft sensors enable it to have better controllability in real problems. The multifunctionality of this actuator has been validated with several experimental tests, and also we have shown it has excellent potential in gripping a variety of objects. Finally, the embedded sensors can discriminate the main functions of actuators, and also they can play the role of independent sensors as well like a stretch, pressure, or bending sensors.

Key words— Soft Robotics, Soft Actuator, Soft Sensor, Soft Gripper, Multifunctional, Kinaesthesia.

## I. INTRODUCTION

Soft robotics is a branch of robotics that uses flexible and soft materials instead of rigid materials[1]. This class of robots is expected to interact with the environment more efficiently, safer, more flexible, and with a higher degree of freedom. There is a variety of soft actuators such as electroactive polymer-based, dielectric elastomers, Pneumatic soft actuators, etc. [2-8]. Nature could be an effective inspiration source to develop new ideas, benefit from the living world's natural optimization, and reduce the number of failed trials and errors [9-11]. Thanks to their nonskeletal flexible structures, plants are one of the best bio-inspiration sources for soft robotics. Their organs demonstrate different kinds of deformations such as elongation, expansion, bending, twisting, and coiling to climb [12], grow [13], penetrate to the soil, explore in the soil [14], etc. Regarding their stimuli, different classes of soft actuators have been developed as electrically, magnetically, photo, humidity, thermally responsive, etc.

Pneumatic soft actuators are scalable, affordable, fast working cycle, and insensitive to temperature drift which makes them a demanding type of soft actuators [7, 15]. Pneumatic soft actuators deform by applying the pneumatic pressure, and the direction of this deformation results from their structural configuration. Regarding their structures, they can be classified into two different groups of fiber-reinforced and fiberless soft actuators. In the fiber-reinforced actuators, the fiber, as the stiff component of the actuator, directs the deformation perpendicular to its orientation [16, 17]. In the fiberless actuators, the main body isn't homogeneous, which causes directional deformations [18]. Main deformations in pneumatic soft actuators are bending [19-21], elongation [22], and twisting [23]. Another significant deformation is coiling, a combination of twisting and bending. While fiber-reinforced actuators aren't able to coil employing a single unit actuator [24], several fiberless pneumatic actuators have been developed to generate coiling deformation with one pressure channel [25, 26] or more[27] .

A grooved shell soft actuator with an almost complex internal chamber, causing a challenging fabrication procedure, is developed mimicking the plant tendril [26]. This actuator needs a resistant layer, usually an embedded paper layer, plus the main soft body. Increasing the length of this actuator, more coils will be generated by applying the pressure. Creating a helical internal channel is one of the most straightforward ideas to generate coiling deformation in a cylindrical single-soft material [25]. The number of generated coils depends on the applied pressure, and a maximum of two coils are generated in the actuator.

Despite all of the differences, both of the mentioned works presented actuators have one-way coiling deformation. The idea of a multi-section-multichannel soft pneumatic actuator provides more freedom for generating different deformations [27]. This idea origins from the fact that a noncentral channel causes bending behavior towards the thicker wall by applying pressure. By creating three noncentral channels, different deformations can be generated by applying different pressures in channels. Different sections with independent channels increase the range of deformations that can be generated. In this regard, a three-section actuator needs nine pressure channels, which means a complex three-unit structure with a complex control procedure. Somehow, each section can be considered an independent actuator.

In this work, the idea of creating the helical channel is improved by adding the second helical channel. The actuator does not need a resistant layer, and the geometry of the channels does not complicate the fabrication procedure. Applying different levels of pressure into the channels causes different deformations such as opposite coilings, opposite-bendings, and elongation. So, a single unit soft actuator can generate a wide range of deformations, only with two pressure lines. This capability makes the actuator an outstanding choice to be used as a multifunctional gripper. By embedding two

).

microfluidic channels and injections of a conductive Gel [28] into them, we also can equip the actuator with self-sensing capability or use it as an independent soft sensor.

The rest of this paper consists of three sections. In Section II, the proposed sensorized soft gripper is described. After that, in section III, with several experimental tests, we will show that the proposed gripper and the embedded sensors are working properly, and finally in section IV we have a conclusion on the main points of the proposed sensorized soft gripper.

## II. PROPOSED SENSORIZED SOFT GRIPPER

### A. Actuation Design

The main components of the proposed soft sensorized actuator are composed of a main soft body, two U-shape sensors, two soft side sensor beds, two helical channels, free close end, fixed base, two air supplier tubes, and four tin-coated copper electrodes for the sensors. Fig. 1. A demonstrates the 3D view of the developed actuator. The active part of the main body is colored by cyan, and two inactive parts, demonstrated by blue and fabricated by stiffer silicone, are created at both ends. These inactive parts do not deform under the load because of that the top piece is channel-less, and the bottom piece is fixed as the base support. However, the top piece moves freely. Two tubes are inserted into the base connected to channels to apply the load. Two longitudinal side beds are considered for better performance of the sensors shown by magenta. These beds are fabricated with a softer silicone to allow more flexibility and amplify the deformation of the sensor area, and so the responsiveness of the sensors also will be amplified. These soft beds are next to two U-shape sensor microchannels with a 200 μm thickness. One of these channels is demonstrated by black lines in Fig. 1. C. After filling the conductive gel into the sensor microchannels, four electrodes are placed inside them through the base. These electrodes are shown in orange.

Two helical channels, one clockwise and the other counterclockwise, are created inside the active body with opposite radial starting points on the base. The pitch of these helixes is 196 mm, and its diameter is 5.5 mm, while the diameter of each channel hole is 1.85 mm. These channels are demonstrated by green lines in Fig. 1. D. Since these channels play a critical role in the performance, their geometry and position are optimized for better sensing and actuation. Applying the controlled input pressures into these two channels causes a variety of deformations and actuation. In particular, single pressure, imbalanced pressures, and balanced pressures applied in the channels cause bidirectional coilings, opposite bendings, and elongation, respectively, in a single unit actuator.

### B. Sensorization

The straightforward nature of resistive sensors helps us to predict their behavior easily. The main idea behind the embedded sensor of the proposed gripper is based on the electrical resistance measurement of a conductive material when its shape is changing. Theoretically, when we change the geometry and the shape of an electrically conductive component, a change in the level of electrical resistance of this component will happen. To have this resistive soft sensor, we have used an already developed technique[23] to make two microfluidic channels into the main body of our proposed gripper. Now, If we fill the microchannels with a conductive gel, we will have a fluidic conductive component that is bendable and even stretchable enough and can respond to the broad range of mechanical stimuli like a stretch, coiling, bending, pressure, etc as a highly stretchable soft sensor. As depicted in Fig.1 C, we have two U-shaped microfluidic channels that we fill out with a combination of water and propylene glycol as the conductive component. Two measure the resistance of these conductive components, we simply place two tin-coated copper wires into the microchannels (Fig.1).

### C. Fabrication

The primary strategy of the fabrication is casting with 3D printed molds. Different types of silicons, all from Smooth-On, are utilized to produce the actuator. The primary construction material of the actuator is Ecoflex™ 00-50, while two longitudinal tapes, located at opposite radial positions of the external surface, are made of Ecoflex™ 00-30 to improve sensing ability by increasing the deformation of the area. Also, both ends of the actuator are closed by Dragon Skin™ 30 as a relatively tough silicone. The mold consists of two main cylindrical walls printed with stereolithography (SLA) 3D printing. One of them has a hole at its bottom to inject Dragon Skin™ 30. Both of them have internal grooves to create two soft longitudinal tapes. These main parts are shown in. Figure. 2. A.

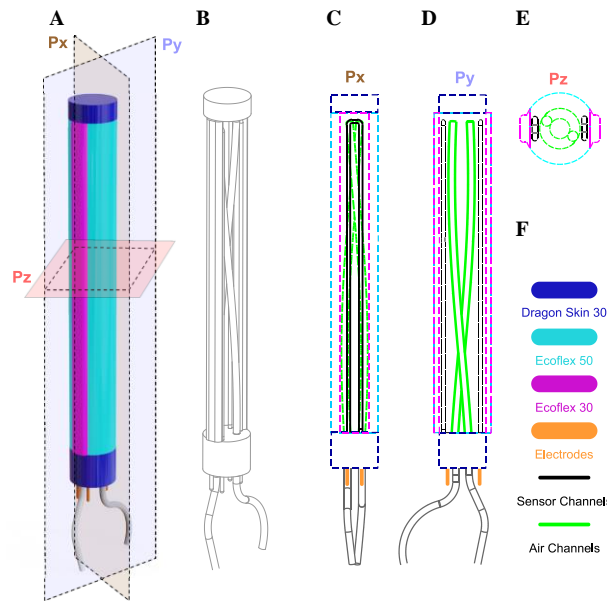

Figure 1. Demonstration of 3D sketch of the sensorized actuator (A). Wireframe display of the same view (B). Section view cutaway by plane Px. Black lines demonstrate one of the U-shape sensor channels (C). Section view cutaway by plane Py. Green lines demonstrate the internal channles, i.e., empty spaces (D). Section view cutaway by plane Pz (E). Each color represents a material (F). Dashed lines demonstrate the cutting edges andnd internal parts

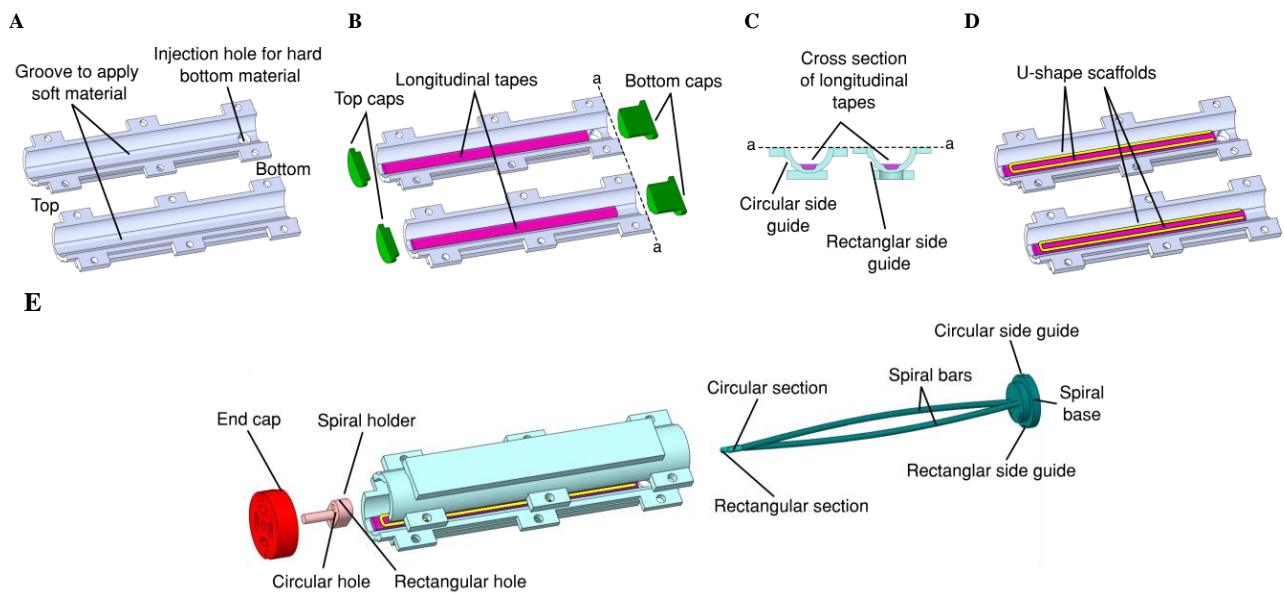

Figure 2. Fabrication procedure of the soft actuator. A) The main mold. B) Using two caps, the softer silicone material is applied to creat two softer longitudinal tapes. C) Cross sectional view of the main molds with longitudinal tapes. D) Two U-shape scaffolds for creating microchannles. E) All of the molding parts in an exploded demonstration.

Fabrication starts with blocking both ends of the main molds with four caps, two caps for each wall, while the main molds are unassembled, as shown in Fig. 2. B. Then, internal grooves are filled by Ecoflex™ 00-30 to create two longitudinal soft tapes in opposite radial locations on the external surface of the actuator. This initial casting results in two soft tapes with the cross-sections shown in Fig. 2. C. Since Ecoflex™ 00-30 is softer than the main body, made by Ecoflex™ 00-50, the actuator deforms so that these tapes experience higher deformations. Since these tapes are used as two beds for the conductive material of sensors, sensorization ability will be more redable. After curing the tapes, caps are removed.

Next, two U-shape scaffolds, made by Acrylonitrile Butadiene Styrene (ABS), are installed on the longitudinal soft tapes, as shown in Fig. 2. D. These scaffolds are 3D printed by the fused deposition modeling (FDM) method with the minimum possible thickness to create two microchannels in the main body close to the external face. ABS filament is used to be sacrificed in acetone after the last molding step.

Then, the main molds are assembled, and helical bars are installed. These bars are printed with SLA connected to their base and are assembled from the bottom side while their top sides are free. Since the bars are thin and may lose the correct related positions, one has a rectangular section top part, and the other one has a circular section top part that is assembled to the related holes on the helical holder. This way, the configuration of the helical related to each other is fixed and matched to the desired design. On the other hand, the position of the helical related to the sensor should be adjusted to improve the sensing ability. In this regard, two side grooves are created on the spiral base and main molds. Aligning these grooves guarantees the correct position of the helical for a better sensing ability (see Fig. 2. E). Now, Dragon Skin™ 30 is injected from the bottom injection hole of the main molds as much as creates a 10 mm stiffer base part. Dragon Skin™ 30 is de-bubbled in a vacuum chamber as long as its pot life after the injection.

After curing the Dragon Skin™ 30, the main body of the actuator is fabricated by injecting Ecoflex™ 00-50 from the top of the mold as much as 10 mm of the top of the main molds remains empty for a better de-bubbling and fabricating the stiff end part. Next, the end cap is installed on the helical holder to centralize the helixes. The end cap and helical holder are printed with the SLA method. Although these two parts could be fabricated in one part, their separation allows easier assembling. Ecoflex™ 00-50 is de-bubbled in a vacuum chamber as long as its pot life after the injection. After curing the Ecoflex™ 00-50, the end cap and helical holder are de-assembled, and the empty part of the end of the mold is filled by Dragon Skin™ 30 to create the stiffer end part. Again, Dragon Skin™ 30 is de-bubbled in a vacuum chamber for as long as its pot life. After curing the end par, all of the molds are de-assembled.

Next, the U-shape scaffolds should be removed by acetone. Since acetone penetrates structures made by silicon without affecting them, it can reach the scaffolds made by ABS and dissolve them. So, the actuator spends over a night in the acetone bath, and then microchannels are washed by acetone to remove residual materials. Creating sensory microchannels, they are filled with conductive materials. These two U-shaped microchannels will be blocked by four tin-coated copper electrodes after injecting the sensory material.

Finally, two tubes are connected to the spiral channels to apply the pressure and generate deformation. Fig1.F demonstrates the 3D view of the final soft actuator with all of its parts.

III. RESULTS AND DISCUSSIONS

A. Actuation

The fabricated actuator has been tested under different loading configurations. The actuator is composed of two internal helical channels, each of which causes coiling in opposite directions. Same as shown in Fig. 3. a,b, applying the pressure just into one channel results in a maximum of two coils, i.e., +720° coiling angle, while applying the pressure into the other channel results in an opposite coiling behavior, i.e., -720° coiling angle.

On the other hand, imbalanced loads generate bending behavior, as shown in Fig. 3. c,d. Also, regarding loading values, pressurizing both channels at the same time causes different deformations. Same loads in both channels cause elongation up to 215%, as shown in Fig. 3. e-i. Concluding, a wide range of different loads can be applied with a wide range of different deformation geometries, which make this actuator a good option to be used as a gripper for different purposes. Fig. 4 demonstrates the ability of this actuator in grasping different objects, from the weight and stiffness of a plant to the weight and strength of a metal tool.

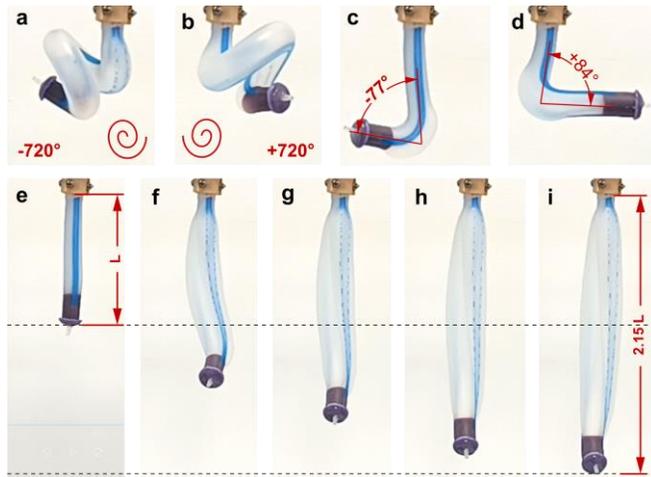

Figure 3. Different deformations of the soft actuator. A) The actuator coils in both opposite directions, i.e., the coiling angle is from -720° to 720°. B) Stimulating both channels with unit air pressure causes an elongation up to about 215%. C) Applying the air pressure to the channels by an imbalanced volume causes bending deformation.

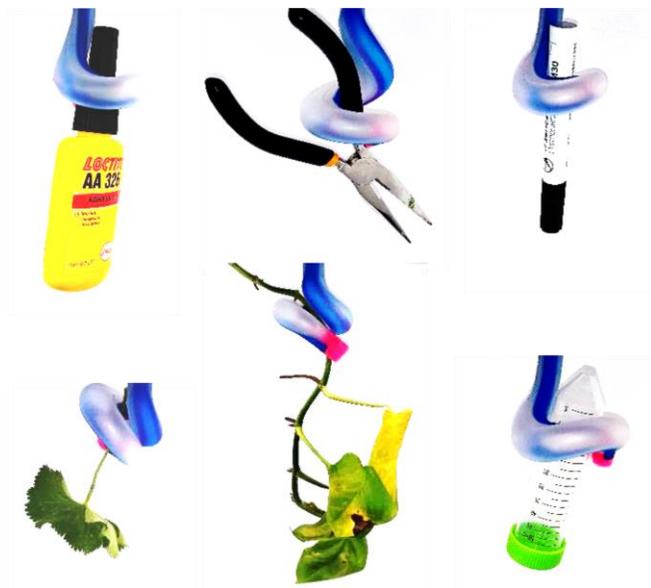

Figure 4. Applying different loads in the channels causes different deformations that make this actuator a good choice for grabbing a wide range of different objects.

## B. Sensing

As mentioned before, the proposed gripper has been equipped with embedded ionic resistive sensors. Here we want to show that these sensors are able to self-sensing of actuation of the gripper, and also the whole system can play the role of an independent soft sensor. To investigate this potentials, some tests have been done. First, we actuate the gripper, and we show that the sensors can discriminate the main actuation of the gripper (Coiling, Bending, and Elongation). Second, we will show that the whole system is responding to an external mechanical stimulus like stretching, bending, etc. Before starting to describe the result, the electronic measurement system should be described.

In this system, as depicted in Fig.5, first, an AC signal is applied to a voltage divider circuit, including the proposed resistive sensor and a constant resistor. Then, due to handling the loading issue, the output of the divider cicuit is buffered and then is fed to a 50Hz Notch filter to remove city power interference. Then, the output is a clean AC signal with no noise. Because the change in the resistance of the sensor would cause the deviate in the amplitude of the AC signal, a peak detector is used to detect the envelope of the signal. Finally, an A/D converter is used to send the analog signal to the computer, and then it is sent to the computer by serial communication. An STM32F103 microcontroller which has a 12-bit A/C was used to convert the analog signal to digital and send it to the computer.

As you can see in Fig.6, the embedded sensors can respond to the stimulated actuations. Hence, by this self-sensing ability, we can understand the state of the actuator needless to have any external sensors like camera, strain gauge, etc. The embedded sensors help the control system also to detect environmental stimulations like blocking by obstacles, oscillation by wind, vibration, and so on. For example, in Fig.7, the gripper has been oscillated by finger movement, and the sensor can generate responses to these disturbances. As another external stimulation, you can see that in response to stretch, the sensors work very synchro (Both signals are very similar to each other), and so we can use this system as an independent soft stretchable sensor. Totally in these examples, we can find that the embedded sensors have enough potential to first self-sensing of the gripper and second play the role of independent soft sensors to respond to the environmental stimulus like oscillation, pressure, stretching, etc.

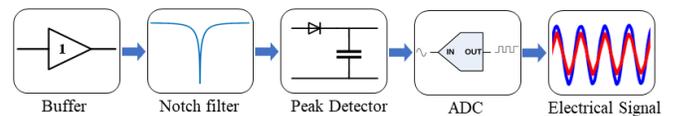

Figure 5. Sensing measurement system.

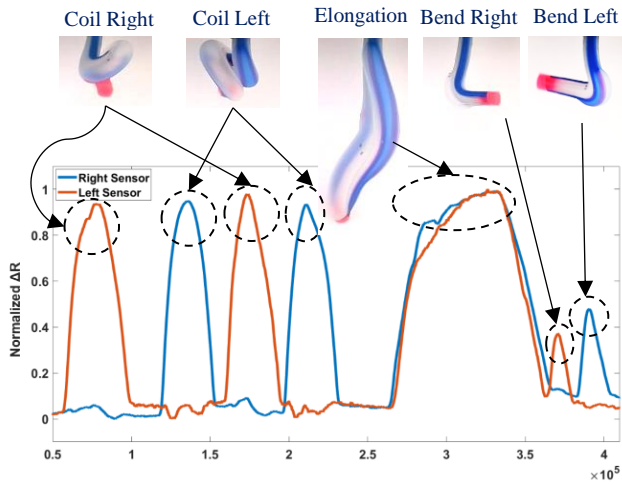

Figure 6. The self-sensing ability of the proposed gripper to discriminate main actuations (Coiling, Bending, and Elongation).

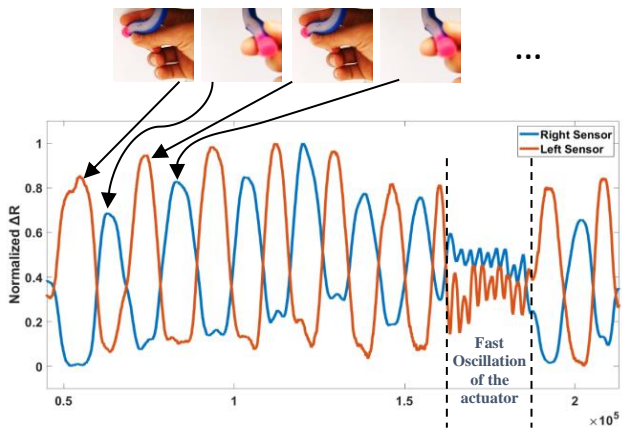

Figure 7. Sensors' responses to environmental stimulations.

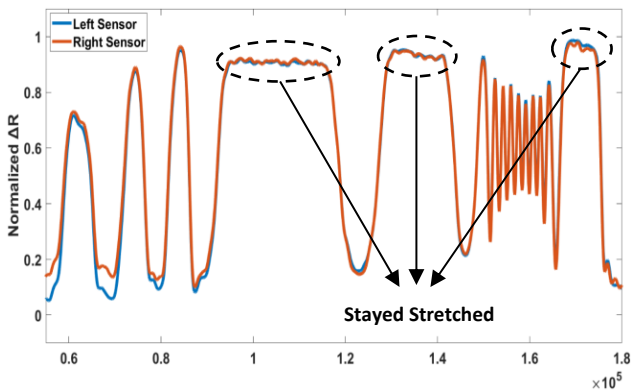

Figure 8. Sensors respond to mechanical stretch as an independent soft sensor.

IV. CONCLUSION

In this paper, we presented a novel fiberless soft pneumatic actuator that can operate as a multifunctional and bidirectional gripper, which is able to bend, coil, and elongate in a single unit actuator. The proposed gripper has been equipped with embedded sensors, which give it a self-proprioception ability. The multifunctionality of this actuator was tested with several experiments showing its potential in a gripping variety of objects from delicate plants to solid metal tools. Some other experimental tests also showed that the embedded sensors are able to discriminate the bending, coiling, and elongation actions of the gripper, and also they can be independent sensors as well to respond to environmental mechanical stimulus like stretch, oscillation, pressure or etc.